\documentclass[10pt,twocolumn,letterpaper]{article}

\usepackage{titling}
\usepackage{cvpr}
\usepackage{times}
\usepackage{epsfig}
\usepackage{graphicx}
\usepackage{amsmath}
\usepackage{amssymb}
\usepackage{gensymb}
\usepackage{multirow}
\usepackage{mathrsfs}

\usepackage[pagebackref=true,breaklinks=true,letterpaper=true,colorlinks,bookmarks=false]{hyperref}

\cvprfinalcopy 

\def\cvprPaperID{3871} 

\ifcvprfinal\fi
\begin{document}

\title{PVNet: Pixel-wise Voting Network for 6DoF Pose Estimation}

\author{
Sida Peng\thanks{Authors contributed equally}\\
Zhejiang University\\
{\tt\small pengsida@zju.edu.cn}
\and
Yuan Liu$^*$\\
Zhejiang University\\
{\tt\small liuyuan@cad.zju.edu.cn}
\and
Qixing Huang\\
UT Austin\\
{\tt\small huangqx@cs.utexas.edu}
\and
Hujun Bao\\
Zhejiang University\\
{\tt\small bao@cad.zju.edu.cn}
\and
Xiaowei Zhou\\
Zhejiang University\\
{\tt\small xzhou@cad.zju.edu.cn}
}

\maketitle

\begin{abstract}
    This paper addresses the challenge of 6DoF pose estimation from a single RGB image under severe occlusion or truncation. Many recent works have shown that a two-stage approach, which first detects keypoints and then solves a Perspective-n-Point (PnP) problem for pose estimation, achieves remarkable performance. However, most of these methods only localize a set of sparse keypoints by regressing their image coordinates or heatmaps, which are sensitive to occlusion and truncation. Instead, we introduce a Pixel-wise Voting Network (PVNet) to regress pixel-wise unit vectors pointing to the keypoints and use these vectors to vote for keypoint locations using RANSAC. This creates a flexible representation for localizing occluded or truncated keypoints. Another important feature of this representation is that it provides uncertainties of keypoint locations that can be further leveraged by the PnP solver. Experiments show that the proposed approach outperforms the state of the art on the LINEMOD, Occlusion LINEMOD and YCB-Video datasets by a large margin, while being efficient for real-time pose estimation. We further create a Truncation LINEMOD dataset to validate the robustness of our approach against truncation. The code will be avaliable at \href{https://zju-3dv.github.io/pvnet/}{https://zju-3dv.github.io/pvnet/}.
\end{abstract}

\section{Introduction}


Object pose estimation aims to detect objects and estimate their orientations and translations relative to a canonical frame~\cite{xiang2014beyond}. Accurate pose estimations are essential for a variety of applications such as augmented reality, autonomous driving and robotic manipulation. For instance, fast and robust pose estimation is crucial in Amazon Picking Challenge~\cite{correll2018analysis}, where a robot needs to pick objects from a warehouse shelf. This paper focuses on the specific setting of recovering the 6DoF pose of an object, i.e., rotation and translation in 3D, from a single RGB image of the object. This problem is quite challenging from many perspectives, including object detection under severe occlusions, variations in lighting and appearance, and cluttered background objects.


Traditional methods~\cite{lowe1999object,lepetit2005monocular,hinterstoisser2012model} have shown that pose estimation can be achieved by establishing the correspondences between an object image and the object model. They rely on hand-crafted features, which are not robust to image variations and background clutters. Deep learning based methods~\cite{su2015render, kehl2017ssd, xiang2017posecnn, bui2018regression} train end-to-end neural networks that take an image as input and output its corresponding pose. However, generalization remains as an issue, as it is unclear that such end-to-end methods learn sufficient feature representations for pose estimation.  


\begin{figure}
    \centering
    \includegraphics[width=1.0\linewidth]{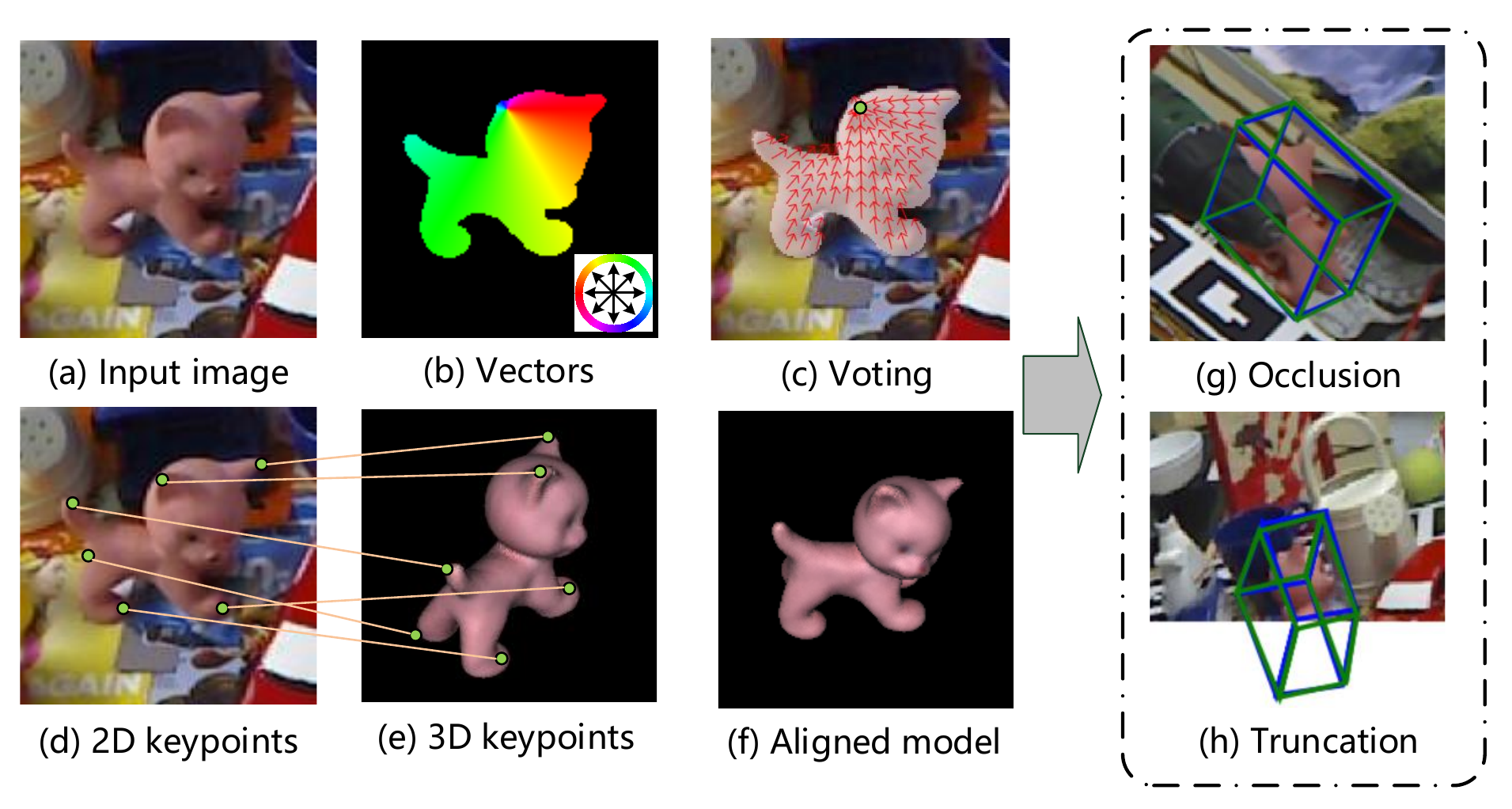}
    \caption{The 6D pose estimation problem is formulated as a Perspective-n-Point (PnP) problem in this paper, which requires correspondences between 2D and 3D keypoints, as illustrated in (d) and (e). We predict unit vectors pointing to keypoints for each pixel, as shown in (b), and localize 2D keypoints in a RANSAC-based voting scheme, as shown in (c). The proposed method is robust to occlusion (g) and truncation (h), where the green bounding boxes represent the ground truth poses and the blue bounding boxes represent our predictions.}
    \label{fig:introduction}
    \vspace{-4mm}
\end{figure}

Some recent methods~\cite{pavlakos20176,rad2017bb8,tekin2018real} use CNNs to first regress 2D keypoints and then compute 6D pose parameters using the Perspective-n-Point (PnP) algorithm. In other words, the detected keypoints serve as an intermediate representation for pose estimation. Such two-stage approaches achieve state-of-the-art performance, thanks to robust detection of keypoints. However, these methods have difficulty in tackling occluded and truncated objects, since part of their keypoints are unseen. Although CNNs may predict these unseen keypoints by memorizing similar patterns, generalization remains difficult. 


We argue that addressing occlusion and truncation requires dense predictions, namely pixel-wise or patch-wise estimates for the final output or intermediate representations. To this end, we propose a novel framework for 6D pose estimation using a Pixel-wise Voting Network (PVNet). The basic idea is illustrated in Figure~\ref{fig:introduction}. Instead of directly regressing image coordinates of keypoints, PVNet predicts unit vectors that represent directions from each pixel of the object towards the keypoints. These directions then vote for the keypoint locations based on RANSAC \cite{fischler1981random}. This voting scheme is motivated from a property of rigid objects that once we see some local parts, we are able to infer the relative directions to other parts. 

Our approach essentially creates a vector-field representation for keypoint localization. In contrast to coordinate or heatmap based representations, learning such a representation enforces the network to focus on local features of objects and spatial relations between object parts. As a result, the location of an invisible part can be inferred from the visible parts. In addition, this vector-field representation is able to represent object keypoints that are even outside the input image. All these advantages make it an ideal representation for occluded or truncated objects. Xiang et al.~\cite{xiang2017posecnn} proposed a similar idea to detect objects and here we use it to localize keypoints. 

Another advantage of the proposed approach is that the dense outputs provide rich information for the PnP solver to deal with inaccurate keypoint predictions. Specifically, RANSAC-based voting prunes outlier predictions and also gives a spatial probability distribution for each keypoint. Such uncertainties of keypoint locations give the PnP solver more freedom to identify consistent correspondences for predicting the final pose. Experiments show that the uncertainty-driven PnP algorithm improves the accuracy of pose estimation.



We evaluate our approach on  LINEMOD~\cite{hinterstoisser2012model}, Occlusion LINEMOD~\cite{brachmann2014learning} and YCB-Video~\cite{xiang2017posecnn} datasets, which are widely-used benchmark datasets for 6D pose estimation. Across all datasets, PVNet exhibits state-of-the-art performances. We also demonstrate the capability of our approach to handle truncated objects on a new dataset called Truncation LINEMOD which is created by randomly cropping images of LINEMOD. Furthermore, our approach is efficient, which runs 25 fps on a GTX 1080ti GPU, to be used for real-time pose estimation.


In summary, this work has the following contributions:
\begin{itemize}
\item We propose a novel framework for 6D pose estimation using a pixel-wise voting network (PVNet), which learns a vector-field representation for robust 2D keypoint localization and naturally deals with occlusion and truncation.
\item We propose to utilize an uncertainty-driven PnP algorithm to account for uncertainties in 2D keypoint localizations, based on the dense predictions from PVNet. 
\item We demonstrate significant performance improvements of our approach compared to the state of the art on benchmark datasets  (ADD: 86.3\% vs. 79\% and 40.8\% vs. 30.4\% on LINEMOD and OCCLUSION, respectively). We also create a new dataset for evaluation on truncated objects.
\end{itemize}

\begin{figure*}
\begin{center}
\includegraphics[width=0.9\textwidth]{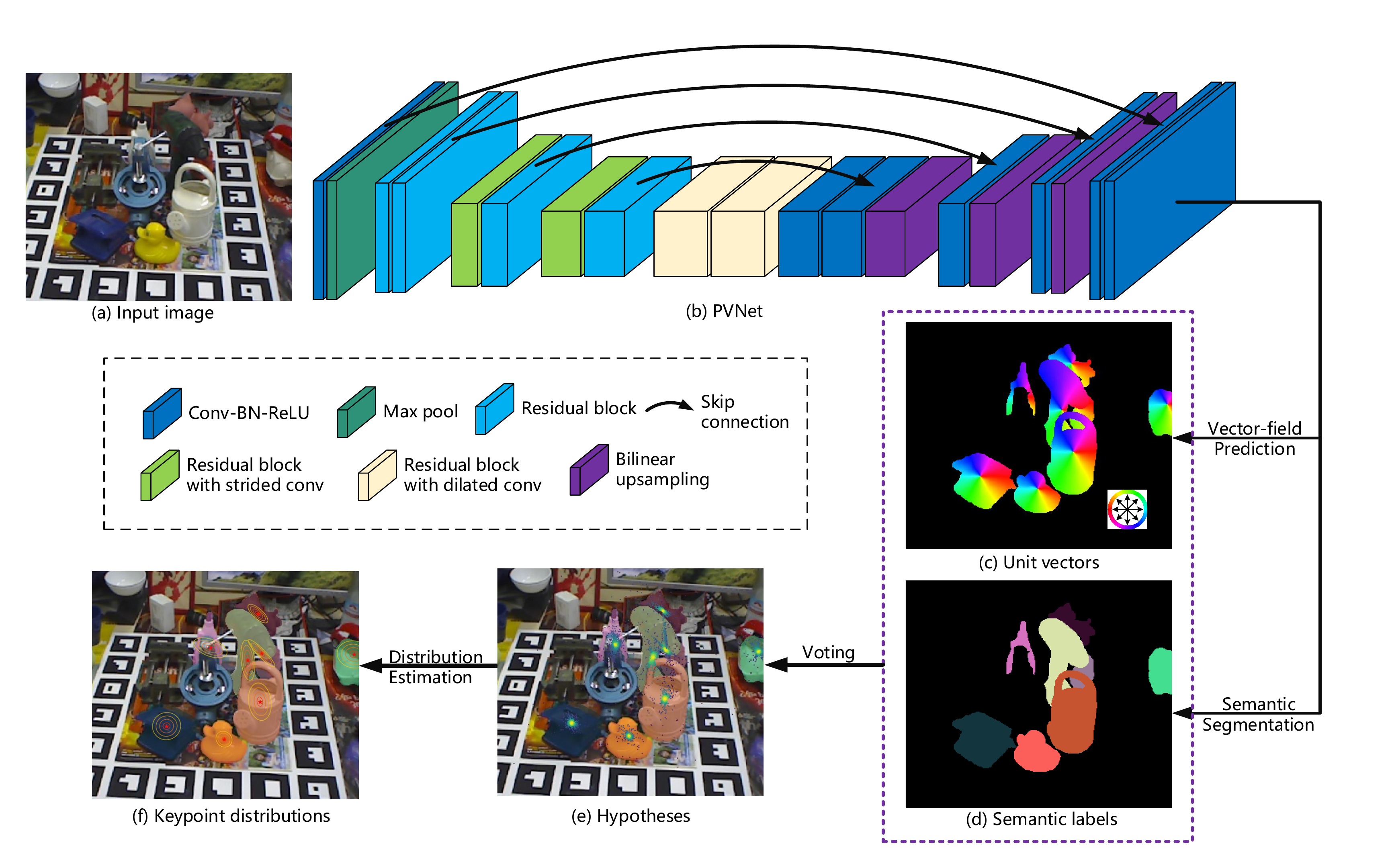}
   \caption{Overview of the keypoint localization: (a) An image of the Occlusion LINEMOD dataset. (b) The architecture of PVNet. (c) Pixel-wise unit vectors pointing to the object keypoints. (d) Semantic labels. (e) Hypotheses of the keypoint locations generated by voting. The hypotheses with higher voting scores are brighter.  (f) Probability distributions of the keypoint locations estimated from hypotheses. The mean of a distribution is represented by a red star and the covariance matrix is shown by ellipses.}
    \label{fig:pipeline}
\end{center}
\vspace{-4mm}
\end{figure*}

\section{Related work}

\paragraph{Holistic methods.} Given an image, some methods aim to estimate the 3D location and orientation of the object in a single shot. Traditional methods mainly rely on template matching techniques~\cite{huttenlocher1993comparing, gu2010discriminative, hinterstoisser2012gradient, zhu2014single}, which are sensitive to cluttered environments and appearance changes. Recently, CNNs have shown significant robustness to environment variations. As a pioneer, PoseNet~\cite{kendall2015posenet} introduces a CNN architecture to directly regress a 6D camera pose from a single RGB image, a task similar to object pose estimation. However, directly localizing objects in 3D is difficult due to a lack of depth information and the large search space. To overcome this problem, PoseCNN~\cite{xiang2017posecnn} localizes objects in the 2D image and predicts their depths to obtain the 3D location. However, directly estimating the 3D rotation is also difficult, since the non-linearity of the rotation space makes CNNs less generalizable. To avoid this problem,~\cite{tulsiani2015viewpoints, su2015render, liu2016ssd, sundermeyer2018implicit} discretize the rotation space and cast the 3D rotation estimation into a classification task. Such discretization produces a coarse result and a post-refinement is essential to get an accurate 6DoF pose.

\paragraph{Keypoint-based methods.} Instead of directly obtaining the pose from an image, keypoint-based methods adopt a two-stage pipeline: they first predict 2D keypoints of the object and then compute the pose through 2D-3D correspondences with a PnP algorithm. 2D keypoint detection is relatively easier than 3D localization and rotation estimation. For objects of rich textures, traditional methods~\cite{lowe1999object, rothganger20063d, bay2006surf} detect local keypoints robustly, so the object pose is estimated both efficiently and accurately, even under cluttered scenes and severe occlusions. However, traditional methods have difficulty in handling texture-less objects and processing low-resolution images~\cite{lepetit2005monocular}. To solve this problem, recent works define a set of semantic keypoints and use CNNs as keypoint detectors.~\cite{rad2017bb8} uses segmentation to identify image regions that contain objects and regresses keypoints from the detected image regions.~\cite{tekin2018real} employs the YOLO architecture~\cite{redmon2017yolo9000} to estimate the object keypoints. Their networks make predictions based on a low-resolution feature map. When global distractions occur, such as occlusions, the feature map is interfered~\cite{oberweger2018making} and the pose estimation accuracy drops. Motivated by the success of 2D human pose estimation~\cite{newell2016stacked}, another category of methods~\cite{pavlakos20176, oberweger2018making} outputs pixel-wise heatmaps of keypoints to address the issue of occlusion. However, since heatmaps are fix-sized, these methods have difficulty in handling truncated objects, whose keypoints may be outside the input image. In contrast, our method makes pixel-wise predictions for 2D keypoints using a more flexible representation, i.e., vector field. The keypoint locations are determined by voting from the directions, which are suitable for truncated objects.

\paragraph{Dense methods.} In these methods, every pixel or patch produces a prediction for the desired output, and then casts a vote for the final result in a generalized Hough voting scheme \cite{liebelt2008independent, sun2010depth, glasner2011aware}. \cite{brachmann2014learning, michel2017global} use a random forest to predict 3D object coordinates for each pixel and produce 2D-3D correspondence hypotheses using geometric constraints. To utilize the powerful CNNs, \cite{kehl2016deep, doumanoglou2016recovering} densely sample image patches and use networks to extract features for the latter voting. 
However, these methods require RGB-D data. 
In the presence of RGB data alone, \cite{brachmann2016uncertainty} uses an auto-context regression framework \cite{tu2010auto} to produce pixel-wise distributions of 3D object coordinates. 
Compared with sparse keypoints, object coordinates provide dense 2D-3D correspondences for pose estimation, which is more robust to occlusion. But regressing object coordinates is more difficult than keypoint detection due to the larger output space. Our approach makes dense predictions for keypoint localization. It can be regarded as a hyprid of keypoint-based and dense methods, which combines advantages of both methods.

\section{Proposed approach}


In this paper, we propose a novel framework for 6DoF object pose estimation. Given an image, the task of pose estimation is to detect objects and estimate their orientations and translations in 3D. Specifically, 6D pose is represented by a rigid transformation $(R; {\bf t})$ from the object coordinate system to the camera coordinate system, where $R$ represents the 3D rotation and $\bf t$ represents the 3D translation.

Inspired by recent methods~\cite{pavlakos20176, rad2017bb8, tekin2018real}, we estimate the object pose using a two-stage pipeline: we first detect 2D object keypoints using CNNs and then compute 6D pose parameters using the PnP algorithm. Our innovation is in a new representation for 2D object keypoints as well as a modified PnP algorithm for pose estimation. Specifically, our method uses a Pixel-wise Voting Network (PVNet) to detect 2D keypoints in a RANSAC-like fashion, which robustly handles occluded and truncated objects. The RANSAC-based voting also gives a spatial probability distribution of each keypoint, allowing us to estimate the 6D pose with an uncertainty-driven PnP.


\subsection{Voting-based keypoint localization}
\label{sec:pvnet}


Figure~\ref{fig:pipeline} overviews the proposed pipeline for keypoint localization. Given an RGB image, PVNet predicts pixel-wise object labels and unit vectors that represent the direction from every pixel to every keypoint. Given the directions to a certain object keypoint from all pixels belonging to that object, we generate hypotheses of 2D locations for that keypoint as well as the confidence scores through RANSAC-based voting. Based on these hypotheses, we estimate the mean and covariance of the spatial probability distribution for each keypoint.

In contrast to directly regressing keypoint locations from an image patch~\cite{rad2017bb8, tekin2018real}, the task of predicting pixel-wise directions enforces the network to focus more on local features of objects and alleviates the influence of cluttered background. Another advantage of this approach is the ability to represent keypoints that are occluded or outside the image. Even if a keypoint is invisible, it can be correctly located according to the directions estimated from other visible parts of the object.




More specifically, PVNet performs two tasks: semantic segmentation and vector-field prediction. For a pixel ${\bf p}$, PVNet outputs the semantic label that associates it with a specific object and the unit vector ${\bf v}_k({\bf p})$ that represents the direction from the pixel $\bf p$ to a 2D keypoint ${\bf x}_k$ of the object. The vector ${\bf v}_k({\bf p})$ is defined as

\begin{equation}
    {\bf v}_k({\bf p}) =  \frac{{\bf x}_k - {\bf p}}{\|{\bf x}_k - {\bf p}\|_2}.
\end{equation}


Given semantic labels and unit vectors, we generate keypoint hypotheses in a RANSAC-based voting scheme. First, we find the pixels of the target object using semantic labels. Then, we randomly choose two pixels and take the intersection of their vectors as a hypothesis ${\bf h}_{k,i}$ for the keypoint ${\bf x}_k$. This step is repeated $N$ times to generate a set of hypotheses $\{{\bf h}_{k,i} |i = 1,2,...,N\}$ that represent possible keypoint locations. Finally, all pixels of the object vote for these hypotheses. Specifically, the voting score $w_{k,i}$ of a hypothesis ${\bf h}_{k,i}$ is defined as

\begin{equation}
	w_{k,i} = \sum\limits_{{\bf p}\in O} \mathbb{I} \left( \frac{({\bf h}_{k,i} - {\bf p})^T}{\|{\bf h}_{k,i} - {\bf p}\|_2} {\bf v}_k({\bf p}) \ge \theta \right),
\end{equation}
where $\mathbb{I}$ represents the indicator function, $\theta$ is a threshold (0.99 in all experiments), and ${\bf p} \in O$ means that the pixel $\bf p$ belongs to the object $O$. Intuitively, a higher voting score means that a hypothesis is more confident as it coincides with more predicted directions.

The resulting hypotheses characterize the spatial probability distribution of a keypoint in the image. Figure~\ref{fig:pipeline}(e) shows an example.  
Finally, the mean ${\boldsymbol \mu_k}$ and the covariance ${\bf \Sigma}_k$ for a keypoint ${\bf x}_k$ are estimated by:

\begin{equation}
	{\bf \boldsymbol\mu}_k = \frac{ \sum_{i=1}^{N} w_{k,i}{\bf h}_{k,i}}{\sum_{i=1}^{N} w_{k,i}},
\end{equation}

\begin{equation}
	{\bf \Sigma}_k = \frac{\sum_{i=1}^{N} w_{k,i}({\bf h}_{k,i} - {\bf \boldsymbol\mu}_k)({\bf h}_{k,i} - {\bf \boldsymbol\mu}_k)^T}{\sum_{i=1}^{N} w_{k,i}},
\end{equation}
which are used latter for uncertainty-driven PnP described in Section~\ref{sec:uncertainty-driven pnp}.


\paragraph{Keypoint selection.}
\label{sec:keypoint selection}

\begin{figure} 
  \centering 
  \scalebox{0.8}{
  \begin{tabular}{ccc}
    \includegraphics[width=0.3\linewidth]{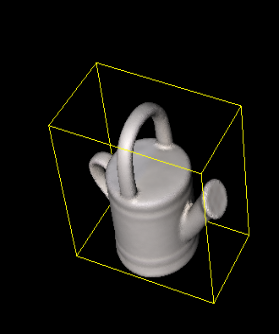} &
    \includegraphics[width=0.3\linewidth]{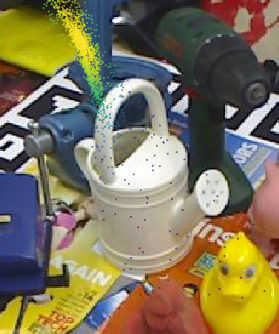}&
    \includegraphics[width=0.3\linewidth]{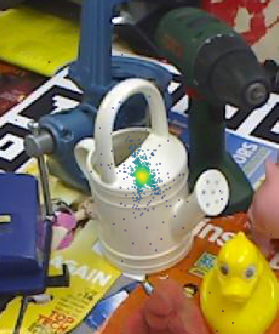}\\
    (a) & (b) & (c) 
  \end{tabular}
  }
  \vspace{2mm}
  \caption{(a) A 3D object model and its 3D bounding box. (b) Hypotheses produced by PVNet for a bounding box corner. (c) Hypotheses produced by PVNet for a keypoint selected on the object surface. The smaller variance of the surface keypoint shows that it is easier to localize the surface keypoint than the bounding box corner in our approach.
  }
  \label{fig:keypoints_selection}
  \vspace{-4mm}
\end{figure}

The keypoints need to be defined based on the 3D object model. Many recent methods~\cite{rad2017bb8, tekin2018real, oberweger2018making} use the eight corners of the 3D bounding box of the object as the keypoints. An example is shown in Figure~\ref{fig:keypoints_selection}(a). These bounding box corners are far away from the object pixels in the image. 
The longer distance to the object pixels results in larger localization errors, since the keypoint hypotheses are generated using the vectors that start at the object pixels.
Figure~\ref{fig:keypoints_selection}(b) and (c) show the hypotheses of a bounding box corner and a keypoint selected on the object surface, respectively, which are generated by our PVNet. The keypoint on the object surface usually has a much smaller variance in the localization. 


Therefore, the keypoints should be selected on the object surface in our approach. Meanwhile, these keypoints should spread out on the object to make the PnP algorithm more stable. Considering the two requirements, we select $K$ kepoints using the farthest point sampling (FPS) algorithm. First, we initialize the keypoint set by adding the object center. Then, we repeatedly find a point on the object surface, which is farthest to the current keypoint set, and add it to the set until the size of the set reaches $K$. The empirical results in Section~\ref{sec:ablation studies} show that this strategy produces better results than using the bounding box corners. We also compare the results using different numbers of keypoints. Considering both accuracy and efficiency, we suggest $K=8$ according to the experiment results.


\paragraph{Multiple instances.}

Our method can handle multiple instances based on the strategy proposed in \cite{xiang2017posecnn, papandreou2018personlab}. For each object class, we generate the hypotheses of the object centers and their voting scores using our proposed voting scheme.
Then, we find the modes among the hypotheses and mark these modes as centers of different instances. Finally, the instance masks are obtained by assigning pixels to the nearest instance center they vote for.


\subsection{Uncertainty-driven PnP}
\label{sec:uncertainty-driven pnp}

Given 2D keypoint locations for each object, its 6D pose can be computed by solving the PnP problem using an off-the-shelf PnP solver, e.g., the EPnP \cite{lepetit2009epnp} used in many previous methods \cite{tekin2018real,rad2017bb8}.
However, most of them ignore the fact that different keypoints may have different confidences and uncertainty patterns, which should be considered when solving the PnP problem.      


As introduced in Section~\ref{sec:pvnet}, our voting-based method estimates a spatial probability distribution for each keypoint. Given the estimated mean ${\boldsymbol \mu_k}$ and covariance matrix ${\bf \Sigma}_k$ for $k=1,\cdots,K$, we compute the 6D pose $(R,{\bf t})$ by minimizing the Mahalanobis distance:

\begin{equation}
\begin{split}
    \underset{R,{\bf t}}{\text{minimize}}~ &\sum\limits_{k=1}^{K} ({\bf \tilde{x}}_k - \boldsymbol\mu_k)^T {\bf \Sigma}_k^{-1} ({\bf \tilde{x}}_k -\boldsymbol\mu_k),
	\\
    &~~{\bf \tilde{x}}_k = \pi(R {\bf X}_k + {\bf t}),
    \label{eq:mdistnace}
\end{split}
\end{equation}
where ${\bf X}_k$ is the 3D coordinate of the keypoint, ${\bf \tilde{x}}_k$ is the 2D projection of ${\bf X}_k$, and $\pi$ is the perspective projection function. The parameters $R$ and $\bf t$ are initialized by EPnP \cite{lepetit2009epnp} based on four keypoints, whose covariance matrices have the smallest traces. Then, we solve \eqref{eq:mdistnace} using the Levenberg-Marquardt algorithm. In \cite{ferraz2014leveraging}, the authors also consider the feature uncertainties by minimizing the approximated Sampson errors. In our method, we directly minimize the reprojection errors.

\section{Implementation details}
Assuming there are $C$ classes of objects and $K$ keypoints for each class, PVNet takes as input the $H\times W\times 3$ image, processes it with a fully convolutional architecture, and outputs the $H\times W\times (K\times 2\times C)$ tensor representing unit vectors and $H\times W \times (C+1)$ tensor representing class probabilities. We use a pretrained ResNet-18 \cite{he2016deep} as the backbone network, and we make three revisions on it. First, when the feature map of the network has the size $H/8\times W/8$, we do not downsample the feature map anymore by discarding the subsequent pooling layers. Second, to keep the receptive fields unchanged, the subsequent convolutions are replaced with suitable dilated convolutions \cite{YuKoltun2016}. Third, the fully connected layers in the original ResNet-18 are replaced with convolution layers. 
Then, we repeatedly perform skip connection, convolution and upsampling on the feature map, until its size reaches $H \times W$, as shown in Figure~\ref{fig:pipeline}(b). 
By applying a $1 \times 1$ convolution on the final feature map, we obtain the unit vectors and class probabilities.

We implement hypothesis generation, pixel-wise voting and density estimation using CUDA. The EPnP \cite{lepetit2009epnp} used to initialize the pose is implemented in OpenCV \cite{bradski2000opencv}. To obtain the final pose, we use the iterative solver Ceres \cite{ceres-solver} to minimize the Mahalanobis distance \eqref{eq:mdistnace}. For symmetric objects, there are ambiguities of keypoint locations. To eliminate the ambiguities, we rotate the symmetric object to a canonical pose during training, as suggested by \cite{rad2017bb8}.

\subsection{Training strategy}

We use the smooth $\ell_1$ loss proposed in \cite{girshick2015fast} for learning unit vectors. The corresponding loss function is defined as
\begin{align}
    \ell({\bf w})=\sum\limits_{k=1}^K &\sum\limits_{{\bf p}\in O}^{} \ell_1({\bf \Delta v}_k({\bf p};{\bf w})|_x) + \ell_1({\bf \Delta v}_k({\bf p};{\bf w})|_y), \nonumber
    \\
    &~~{\bf \Delta v}_k({\bf p};{\bf w})={\bf \tilde{v}}_k({\bf p};{\bf w})-{\bf v}_k({\bf p}),
\end{align}
where $\bf w$ represents the parameters of PVNet, ${\bf \tilde{v}}_k$ is the predicted vector, ${\bf v}_k$ is the ground truth unit vector, and ${\bf \Delta v}_k|_x$ and ${\bf \Delta v}_k|_y$ represent the two elements of ${\bf \Delta v}_k$, respectively. For training semantic labels, a softmax cross-entropy loss is adopted. Note that during testing, we do not need the predicted vectors to be unit because the subsequent processing uses only the directions of the vectors.

To prevent overfitting, we add synthetic images to the training set. For each object, we render 10000 images whose viewpoints are uniformly sampled. We further synthesize another 10000 images using the ``Cut and Paste" strategy proposed in \cite{dwibedi2017cut}. The background of each synthetic image is randomly sampled from SUN397~\cite{xiao2010sun}. We also apply online data augmentation including random cropping, resizing, rotation and color jittering during training. We set the initial learning rate as 0.001 and halve it every 20 epochs. All models are trained for 200 epochs.

\section{Experiments}

\subsection{Datasets}

\paragraph{LINEMOD~\cite{hinterstoisser2012model}} is a standard benchmark for 6D object pose estimation. This dataset exhibits many challenges for pose estimation: cluttered scenes, texture-less objects, and lighting condition variations.

\paragraph{Occlusion LINEMOD~\cite{brachmann2014learning}} was created by additionally annotating a subset of the LINEMOD images. Each image contains multiple annotated objects, and these objects are heavily occluded, which poses a great challenge for pose estimation.

\paragraph{Truncation LINEMOD} To fully evaluate our method on truncated objects, we create this dataset by randomly cropping images in the LINEMOD dataset. After cropping, only 40\% to 60\% of the area of the target object remain in the image. 
Some examples are shown in Figure \ref{fig:trun_result}. 

Note that, in our experiments, the Occlusion LINEMOD and Truncation LINEMOD are used for testing only. Our model tested on these two datasets is only trained on the LINEMOD dataset.

\paragraph{YCB-Video~\cite{xiang2017posecnn}} is a recently proposed dataset. The images are collected from the YCB object set~\cite{calli2015ycb}. This dataset is challenging due to the varying lighting conditions, significant image noise and occlusions.

\subsection{Evalutation metrics}

We evaluate our method using two common metrics: 2D projection metric~\cite{brachmann2016uncertainty} and average 3D distance of model points (ADD) metric~\cite{hinterstoisser2012model}.

\paragraph{2D Projection metric.} This metric computes the mean distance between the projections of 3D model points given the estimated and the ground truth pose. A pose is considered as correct if the distance is less than 5 pixels.

\paragraph{ADD metric.} With the ADD metric~\cite{hinterstoisser2012model}, we transform the model points by the estimated and the ground truth poses, respectively, and compute the mean distance between the two transformed point sets. When the distance is less than 10\% of the model's diameter, it is claimed that the estimated pose is correct. For symmetric objects, we use the ADD-S metric~\cite{xiang2017posecnn}, where the mean distance is computed based on the closest point distance. We denote these two metrics as ADD(-S) and use the one appropriate to the object. When evaluating on the YCB-Video dataset, we compute the ADD(-S) AUC proposed in \cite{xiang2017posecnn}. The ADD(-S) AUC is the area under the accuracy-threshold curve, which is obtained by varying the distance threshold in evaluation.

\subsection{Ablation studies}
\label{sec:ablation studies}

\begin{table}
\begin{center}
\scalebox{0.8}{
\begin{tabular}{|c|cccccc|}
\hline
 \multirow{2}{*}{methods} & Tekin & BBox & FPS & FPS & FPS & FPS 8  \\ 
 & \cite{tekin2018real} & 8 & 4 & 8 & 12 & + Un  \\ \hline
ape & 2.48 & 6.50 & 5.31 & \textbf{17.44} & 15.1 & 15.81 \\ 
can & 17.48 & \textbf{65.04} & 18.81 & 63.21 & 64.87 & 63.30 \\
cat & 0.67 & 15.00 & 16.01 & \textbf{17.35} & 16.68 & 16.68 \\
duck & 1.14 & 15.95 & 13.85 & \textbf{26.12} & 24.89 & 25.24 \\
driller & 7.66 & 55.60 & 12.19 & 62.19 & 64.17 & \textbf{65.65} \\
eggbox & - & 35.23 & 36.77 & 44.96 & 41.53 & \textbf{50.17} \\
glue & 10.08 & 42.64 & 24.81 & 47.32 & \textbf{51.94} & 49.62 \\
holepuncher & 5.45 & 35.06 & 15.98 & 39.50 & \textbf{40.16} & 39.67 \\\hline
average & 6.42 & 33.88 & 17.96 & 39.76 & 39.92 & \textbf{40.77} \\ \hline
\end{tabular}
} 
\vspace{-0.2mm}
\end{center}
\caption{Ablation studies on different configurations for pose estimation on the \textbf{Occlusion LINEMOD} dataset. These results are accuracies in terms of the \textbf{ADD(-S)} metric, where glue and eggbox are considered as symmetric objects. Tekin \cite{tekin2018real} detects the keypoints by regression, while other configurations use the proposed voting-based keypoint localization. \textbf{BBox 8} shows the result of our method using the keypoints defined in \cite{tekin2018real}. \textbf{FPS $\textbf{K}$} means that we detect $K$ surface keypoints generated by the FPS algorithm. \textbf{Un} means that we use the uncertainty-driven PnP. In configurations without \textbf{Un}, the pose is estimated using the EPnP \cite{lepetit2009epnp}.}
\vspace{-2mm}
\label{tab:ablation}
\end{table}


We conduct ablation studies to compare different keypoint detection methods, keypoint selection schemes, numbers of keypoints and PnP algorithms, on the Occlusion LINEMOD dataset. Table~\ref{tab:ablation} summarizes the results of ablation studies. 

To compare PVNet with \cite{tekin2018real}, we re-implement the same pipeline as \cite{tekin2018real} but use PVNet to detect the keypoints which include 8 bounding box corners and the object center. The result is listed in the column ``BBox 8" in Table~\ref{tab:ablation}. The column ``Tekin" shows the original result of~\cite{tekin2018real}, which directly regresses coordinates of keypoints via a CNN. Comparing the two columns demonstrates that pixel-wise voting is more robust to occlusion.

To analyze the keypoint selection schemes discussed in Section~\ref{sec:keypoint selection}, we compare the pose estimation results based on different keypoint sets: ``BBox 8" that includes 8 bounding box corners plus the center and  ``FPS 8" that includes 8 surface points selected by the FPS algorithm plus the center. Comparing ``BBox 8" with ``FPS 8" in Table~\ref{tab:ablation} shows that the proposed FPS scheme results in better pose estimation.

When exploring the influence of the keypoint number on pose estimation, we train PVNet to detect 4, 8 and 12 surface keypoints plus the object center, respectively. All the three sets of keypoints are selected by the FPS algorithm as described in Section~\ref{sec:keypoint selection}. Comparing columns ``FPS 4", ``FPS 8" and ``FPS 12" shows that the accuracy of pose estimation increases with the keypoint number. But the gap between ``FPS 8" and ``FPS 12" is negligible. Considering efficiency, we use ``FPS 8" in all the other experiments. 

To validate the benefit of considering the uncertainties in solving the PnP problem, we replace the EPnP~\cite{lepetit2009epnp} used in ``FPS 8" with the uncertainty-driven PnP. The results are shown in the last column ``FPS 8 + Un" in Table~\ref{tab:ablation}, which demonstrate that considering uncertainties of keypoint locations improves the accuracy of pose estimation.

The configuration ``FPS 8 + Un" is the final configuration for our approach, which is denoted by ``OURS" in the following experiments.

\begin{table}
\begin{center}
\scalebox{0.8}{
\begin{tabular}{|c|ccc|c|}
\hline
 & \multicolumn{3}{c|}{w/o refinement} & w/ refinement \\\hline
\multirow{2}{*}{methods} & BB8 & Tekin & OURS & BB8 \\
 &\cite{rad2017bb8} & \cite{tekin2018real} & & \cite{rad2017bb8}\\ \hline
ape & 95.3 & 92.10 & \textbf{99.23} & 96.6 \\
benchwise & 80.0 & 95.06 & \textbf{99.81} & 90.1 \\
cam & 80.9 & 93.24 & \textbf{99.21} & 86.0 \\ 
can & 84.1 & 97.44 & \textbf{99.90} & 91.2 \\ 
cat & 97.0 & 97.41 & \textbf{99.30} & 98.8 \\ 
driller & 74.1 & 79.41 & \textbf{96.92} & 80.9 \\ 
duck & 81.2 & 94.65 & \textbf{98.02} & 92.2 \\ 
eggbox & 87.9 & 90.33 & \textbf{99.34} & 91.0 \\ 
glue & 89.0 & 96.53 & \textbf{98.45} & 92.3 \\ 
holepuncher & 90.5 & 92.86 & \textbf{100.0} & 95.3 \\ 
iron & 78.9 & 82.94 & \textbf{99.18} & 84.8 \\ 
lamp & 74.4 & 76.87 & \textbf{98.27} & 75.8 \\ 
phone & 77.6 & 86.07 & \textbf{99.42} & 85.3 \\ \hline
average & 83.9 & 90.37 & \textbf{99.00} & 89.3 \\ \hline
\end{tabular}
} 
\end{center}
	\vspace{-0.2mm}
\caption{The accuracies of our method and the baseline methods on the \textbf{LINEMOD} dataset in terms of the \textbf{2D projection} metric. }
	\vspace{-2mm}
\label{tab:linemod_proj}
\end{table}

\subsection{Comparison with the state-of-the-art methods}

\begin{table}
\tabcolsep=0.07cm
\begin{center}

\scalebox{0.8}{
\begin{tabular}{|c|cccc|cc|}
\hline
  & \multicolumn{4}{c|}{w/o refinement} & \multicolumn{2}{c|}{w/ refinement}\\ 
\hline

\multirow{2}{*}{methods} & BB8 & SSD-6D & Tekin & OURS & BB8 & SSD-6D\\ 
 & \cite{rad2017bb8} & \cite{kehl2017ssd} &\cite{tekin2018real} & & \cite{rad2017bb8} & \cite{kehl2017ssd} \\\hline
ape & 27.9 & 0.00 & 21.62 & 43.62 & 40.4 & \textbf{65} \\ 
benchwise & 62.0 & 0.18 & 81.80 & \textbf{99.90} & 91.8 & 80 \\
cam & 40.1 & 0.41 & 36.57 & \textbf{86.86} & 55.7 & 78 \\ 
can & 48.1 & 1.35 & 68.80 & \textbf{95.47} & 64.1 & 86 \\ 
cat & 45.2 & 0.51 & 41.82 & \textbf{79.34} & 62.6 & 70 \\ 
driller & 58.6 & 2.58 & 63.51 & \textbf{96.43} & 74.4 & 73 \\ 
duck & 32.8 & 0.00 & 27.23 & 52.58 & 44.30 & \textbf{66} \\ 
eggbox & 40.0 & 8.90 & 69.58 & 99.15 & 57.8 & \textbf{100} \\ 
glue & 27.0 & 0.00 & 80.02 & 95.66 & 41.2 & \textbf{100} \\ 
holepuncher & 42.4 & 0.30 & 42.63 & \textbf{81.92} & 67.20 & 49 \\ 
iron & 67.0 & 8.86 & 74.97 & \textbf{98.88} & 84.7 & 78 \\ 
lamp & 39.9 & 8.20 & 71.11 & \textbf{99.33} & 76.5 & 73 \\ 
phone & 35.2 & 0.18 & 47.74 & \textbf{92.41} & 54.0 & 79 \\ \hline
average & 43.6 & 2.42 & 55.95 & \textbf{86.27} & 62.7 & 79 \\ \hline
\end{tabular}
} 
\vspace{-0.2mm}
\end{center}
\caption{The accuracies of our method and the baseline methods on the \textbf{LINEMOD} dataset in terms of the \textbf{ADD(-S)} metric, where glue and eggbox are considered as symmetric objects. }
\vspace{-2mm}
\label{tab:linemod_add}
\end{table}

We compare with the state-of-the-art methods which take RGB images as input and output 6D object poses.

\paragraph{Performance on the LINEMOD dataset.} In Table~\ref{tab:linemod_proj}, we compare our method with~\cite{rad2017bb8, tekin2018real} on the LINEMOD dataset in terms of the 2D projection metric. ~\cite{rad2017bb8, tekin2018real} detect keypoints by regression, while our method uses the proposed voting-based keypoint localization. BB8~\cite{rad2017bb8} trains another CNN to refine the predicted pose and the refined results are shown in a separate column. Our method achieves the state-of-the-art performance on all objects without the need of a separate refinement stage.

Table~\ref{tab:linemod_add} shows the comparison of our methods with~\cite{rad2017bb8, liu2016ssd, tekin2018real} in terms of the ADD(-S) metric. Note that we compute the ADD-S metric for the eggbox and the glue, which are symmetric, as suggested in~\cite{xiang2017posecnn}. Comparing to these methods without using refinement, our method outperforms them by a large margin of at least 30.32\%. SSD-6D~\cite{kehl2017ssd} significantly improves its own performance using edge alignment to refine the estimated pose. Nevertheless, our method still outperforms it by 7.27\%.

\begin{table}
\begin{center}
\scalebox{0.8}{
\begin{tabular}{|c|cccc|}
\hline
\multirow{2}{*}{methods} & Tekin & PoseCNN & Oberweger & OURS \\ 
& \cite{tekin2018real} & \cite{xiang2017posecnn} & \cite{oberweger2018making} & \\ \hline
ape & 7.01 & 34.6 & \textbf{69.6} & 69.14 \\ 
can & 11.20 & 15.1 & 82.6 & \textbf{86.09} \\ 
cat & 3.62 & 10.4 & 65.1 & \textbf{65.12} \\ 
duck & 5.07 & 31.8 & 61.4 & \textbf{61.44} \\ 
driller & 1.40 & 7.4 & \textbf{73.8} & 73.06 \\ 
eggbox & - & 1.9 & \textbf{13.1} & 8.43 \\ 
glue & 4.70 & 13.8 & 54.9 & \textbf{55.37} \\ 
holepuncher & 8.26 & 23.1 & 66.4 & \textbf{69.84} \\ \hline
average & 6.16 & 17.2 & 60.9 & \textbf{61.06} \\ \hline
\end{tabular}

} 
\end{center}
	\vspace{-0.2mm}
\caption{The accuracies of our method and the baseline methods on the \textbf{Occlusion LINEMOD} dataset in terms of the \textbf{2D projection} metric. }
	\vspace{-2mm}
\label{tab:occ_prj}
\end{table}

\begin{table}
\begin{center}
\scalebox{0.8}{
\begin{tabular}{|c|cccc|}
\hline
 \multirow{2}{*}{methods} & Tekin & PoseCNN & Oberweger & OURS \\ 
 & \cite{tekin2018real} & \cite{xiang2017posecnn} & \cite{oberweger2018making} & \\ \hline
ape & 2.48 & 9.6 & \textbf{17.6} & 15.81 \\ 
can & 17.48 & 45.2 & 53.9 & \textbf{63.30} \\ 
cat & 0.67 & 0.93 & 3.31 & \textbf{16.68} \\ 
duck & 1.14 & 19.6 & 19.2 & \textbf{25.24} \\ 
driller & 7.66 & 41.4 & 62.4 & \textbf{65.65} \\ 
eggbox & - & 22 & 25.9 & \textbf{50.17} \\ 
glue & 10.08 & 38.5 & 39.6 & \textbf{49.62} \\ 
holepuncher & 5.45 & 22.1 & 21.3 & \textbf{39.67} \\ \hline
average & 6.42 & 24.9 & 30.4 & \textbf{40.77} \\ \hline
\end{tabular}
} 
\end{center}
	\vspace{-0.2mm}
\caption{The accuracies of our method and the baseline methods on the \textbf{Occlusion LINEMOD} dataset in terms of the \textbf{ADD(-S)} metric, where glue and eggbox are considered as symmetric objects.}
	\vspace{-2mm}
\label{tab:occ_add}
\end{table}

\begin{figure*}[t]
	\centering
	\scalebox{0.3}{
		\begin{tabular}{ccccccccc}
			\includegraphics[width=0.4\linewidth]{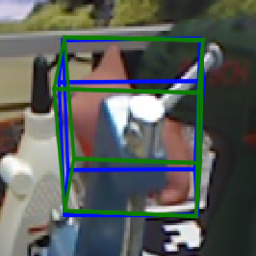}
			&\includegraphics[width=0.4\linewidth]{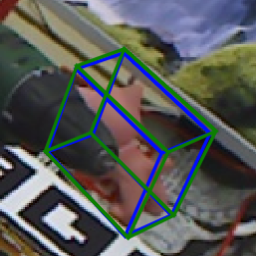}
			&\includegraphics[width=0.4\linewidth]{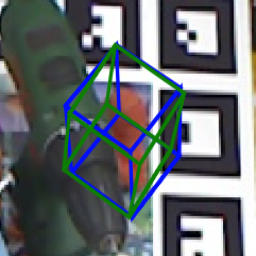}
			&\includegraphics[width=0.4\linewidth]{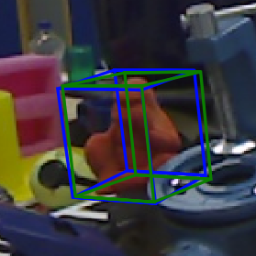}
			&\includegraphics[width=0.4\linewidth]{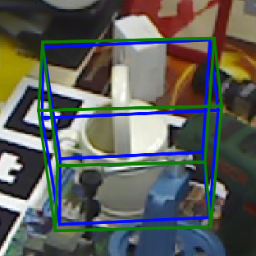}
			&\includegraphics[width=0.4\linewidth]{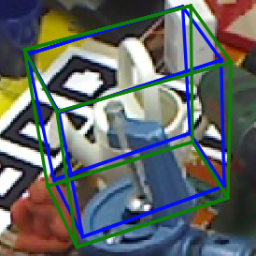}
			&\includegraphics[width=0.4\linewidth]{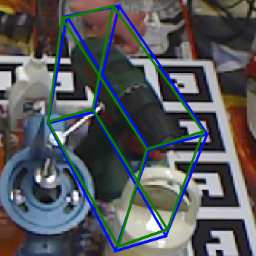}
			&\includegraphics[width=0.4\linewidth]{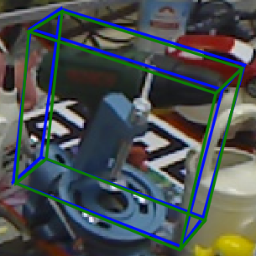}
			\\\includegraphics[width=0.4\linewidth]{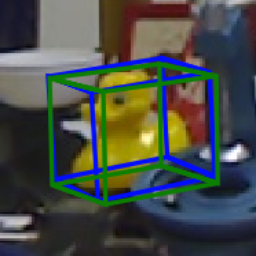}
			&\includegraphics[width=0.4\linewidth]{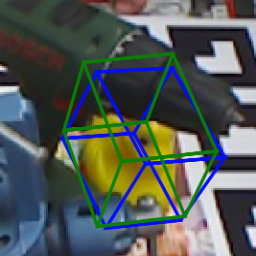}
			&\includegraphics[width=0.4\linewidth]{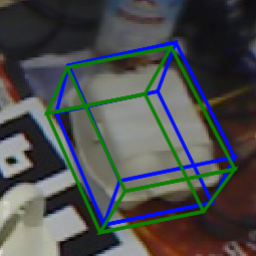}
			&\includegraphics[width=0.4\linewidth]{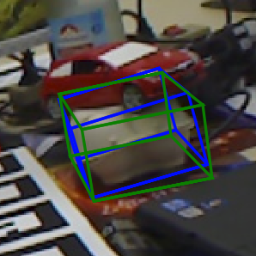}
			&\includegraphics[width=0.4\linewidth]{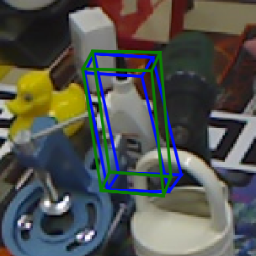}
			&\includegraphics[width=0.4\linewidth]{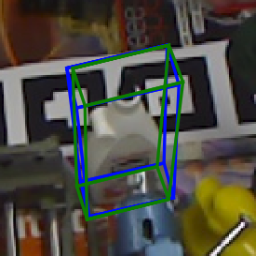}
			&\includegraphics[width=0.4\linewidth]{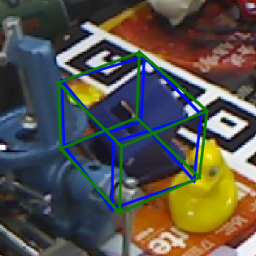}
			&\includegraphics[width=0.4\linewidth]{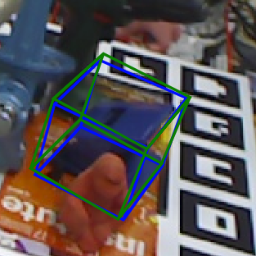}
		\end{tabular}
	} \vspace{1mm}
	\caption{Visualizations of results on the Occlusion LINEMOD dataset. Green 3D bounding boxes represent the ground truth poses while blue 3D bounding boxes represent our predictions.}
	\vspace{-0mm}
	\label{fig:occ_result}
\end{figure*} 
\begin{figure*}[t]
	\centering
	\scalebox{0.3}{
		\begin{tabular}{cccccccc}
			\includegraphics[width=0.4\linewidth]{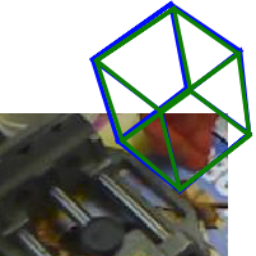} &
			\includegraphics[width=0.4\linewidth]{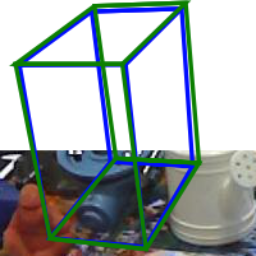} &
			\includegraphics[width=0.4\linewidth]{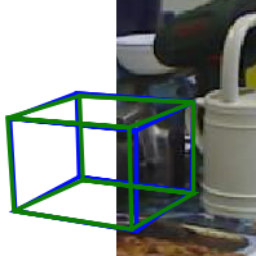} &
			\includegraphics[width=0.4\linewidth]{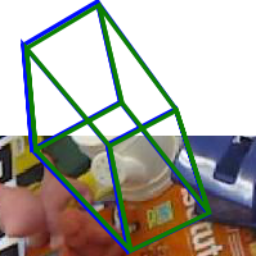} &
			\includegraphics[width=0.4\linewidth]{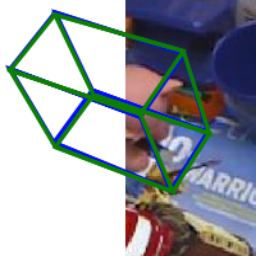} &
			\includegraphics[width=0.4\linewidth]{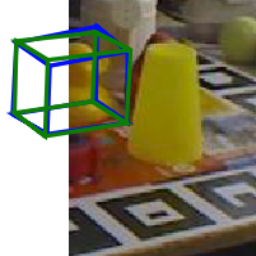} &
			\includegraphics[width=0.4\linewidth]{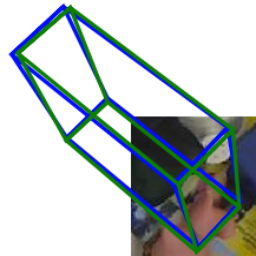} &
			\includegraphics[width=0.4\linewidth]{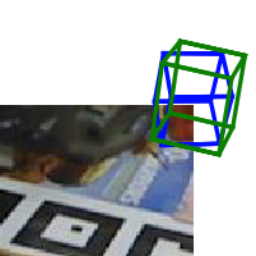} \\
			\includegraphics[width=0.4\linewidth]{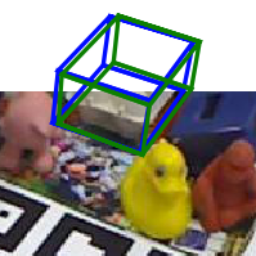} &
			\includegraphics[width=0.4\linewidth]{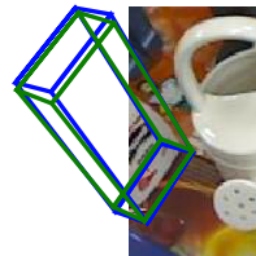} &
			\includegraphics[width=0.4\linewidth]{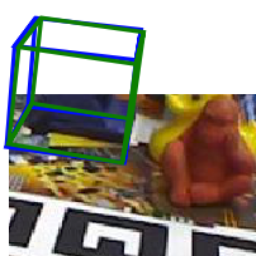} &
			\includegraphics[width=0.4\linewidth]{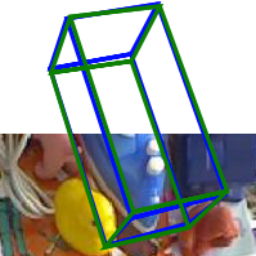} &
			\includegraphics[width=0.4\linewidth]{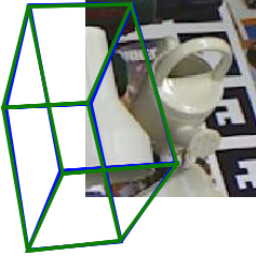} &
			\includegraphics[width=0.4\linewidth]{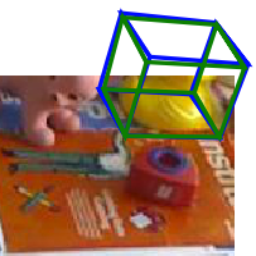} &
			\includegraphics[width=0.4\linewidth]{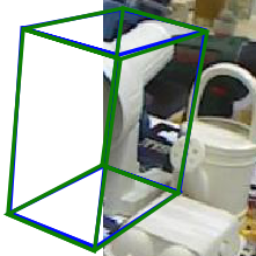} &
			\includegraphics[width=0.4\linewidth]{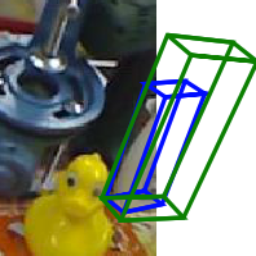}
		\end{tabular}
	} \vspace{2mm}
	\caption{We create a new dataset named Truncation LINEMOD by randomly cropping each image of the LINEMOD dataset. Visualizations of results on the Truncation LINEMOD dataset are shown. Green 3D bounding boxes represent the ground truth poses while blue 3D bounding boxes represent our predictions. The images of the last column are the failure cases, where the visible parts are too ambiguous to provide enough information for the pose estimation.}
	\vspace{-4mm}
	\label{fig:trun_result}
\end{figure*}

\paragraph{Robustness to occlusion.} We use the model trained on the LINEMOD dataset for testing on the Occlusion LINEMOD dataset. Table~\ref{tab:occ_prj} and Table~\ref{tab:occ_add} summarize the comparison with~\cite{tekin2018real, xiang2017posecnn, oberweger2018making} on the Occlusion LINEMOD dataset in terms of the 2D projection metric and the ADD(-S) metric, respectively. For both metrics, our method achieves the best performance among all methods. In particular, our method outperforms other methods by a margin of 10.37\% in terms of the ADD(-S) metric. Some qualitative results are shown in Figure \ref{fig:occ_result}. The improved performance demonstrates that the proposed vector-field representation enables PVNet to learn the relationship between parts of the object, so that the occluded keypoints can be robustly recovered by the visible parts.

\begin{table}
\begin{center}
\scalebox{0.75}
{
\begin{tabular}{|c|ccccccc|}
\hline
 \multirow{2}{*}{objects} & \multirow{2}{*}{ape} & benc- & \multirow{2}{*}{cam}  & \multirow{2}{*}{can} &  \multirow{2}{*}{cat} & \multirow{2}{*}{driller} & \multirow{2}{*}{duck} \\ 
 &  & hvise  & & & & & \\ \hline
2D Projection & 52.59 & 58.19 & 54.87 & 57.44 & 61.66 & 43.27 & 54.23 \\
ADD(-S) & 12.78 & 42.80 & 27.73 & 32.94 & 25.19 & 37.04 & 12.36 \\ \hline
 \multirow{2}{*}{objects} & \multirow{2}{*}{eggbox} & \multirow{2}{*}{glue} & holep- & \multirow{2}{*}{iron} & \multirow{2}{*}{lamp} & \multirow{2}{*}{phone}  & \multirow{2}{*}{avg} \\
 & & & uncher & & & & \\ \hline
2D Projection & 87.23 & 86.64 & 53.84 & 46.53 & 46.94 & 51.35 & 58.06 \\
ADD(-S) & 44.13 & 38.11 & 22.39 & 42.01 & 40.91 & 30.86 & 31.48 \\
\hline
\end{tabular}
}
\end{center}
	\vspace{-0.2mm}
\caption{Our results on the \textbf{Truncation LINEMOD} dataset in terms of the \textbf{2D projection} and the \textbf{ADD(-S)} metrics.}
	\vspace{-4mm}
\label{tab:trun_result}
\end{table}

\paragraph{Robustness to truncation.} We evaluate our method on the Truncation LINEMOD dataset. Note that, the model used for testing is only trained on the LINEMOD dataset. Table~\ref{tab:trun_result} shows quantitative results in terms of the 2D projection and ADD(-S) metrics. We also test the released model from \cite{tekin2018real}, but it does not obtain reasonable results as it is not designed for this case.

Figure~\ref{fig:trun_result} shows some qualitative results. Even the objects are partially visible, our method robustly recovers their poses. We show two failure cases in the last column of Figure~\ref{fig:trun_result}, where the visible parts do not provide enough information to infer the poses. 
This phenomenon is particularly obvious for small objects, such as duck and ape, which have lower accuracies of the pose estimation.

\paragraph{Performance on the YCB-Video dataset.} In Table~\ref{tab:ycb_avg}, we compare our method with~\cite{ xiang2017posecnn, oberweger2018making} on the YCB-Video dataset in terms of the 2D projection and the ADD(-S) AUC metrics. Our method again achieves the state-of-the-art performance and surpasses Oberweger~\cite{oberweger2018making} which is specially designed for dealing with occlusion. The results of PoseCNN were obtained from Oberweger~\cite{oberweger2018making}.

\begin{table}
\begin{center}
\scalebox{0.8}{
\begin{tabular}{|c|ccc|}
\hline
\multirow{2}{*}{methods} & PoseCNN & Oberweger & OURS \\
&\cite{xiang2017posecnn}&\cite{oberweger2018making}& \\ \hline
2D Projection & 3.72 & 39.4 & \textbf{47.4} \\ \hline
ADD(-S) AUC & 61.0 & 72.8 & \textbf{73.4} \\ \hline
\end{tabular}
} 
\end{center}
\vspace{-0.2mm}
\caption{The accuracies of our method and the baseline methods on the YCB-Video dataset in terms of the \textbf{2D projection} and the \textbf{ADD(-S) AUC} metrics.}
\label{tab:ycb_avg}
\vspace{-2mm}
\end{table}

\subsection{Running time}

Given a $480 \times 640$ image, our method runs at 25 fps on a desktop with an Intel i7 3.7GHz CPU and a GTX 1080 Ti GPU, which is efficient for real-time pose estimation. Specifically, our implementation takes 10.9 ms for data loading, 3.3 ms for network forward propagation, 22.8 ms for the RANSAC-based voting scheme, and 3.1 ms for the uncertainty-driven PnP.

\section{Conclusion}

We introduced a novel framework for 6DoF object pose estimation, which consists of the pixel-wise voting network (PVNet) for keypoint localization and the uncertainty-driven PnP for final pose estimation. We showed that predicting the vector fields followed by RANSAC-based voting for keypoint localization gained a superior performance than direct regression of keypoint coordinates, especially for occluded or truncated objects. We also showed that considering the uncertainties of predicted keypoint locations in solving the PnP problem further improved pose estimation. We reported the state-of-the-art performances on all three widely-used benchmark datasets and demonstrated the robustness of the proposed approach on a new dataset of truncated objects.

{\small
\bibliographystyle{ieee}
\bibliography{egbib}
}

\end{document}